# Taxonomy of Pathways to Dangerous AI

Roman V. Yampolskiy

Computer Engineering and Computer Science, Speed School of Engineering, University of Louisville
roman.yampolskiy@louisville.edu

**Abstract**

In order to properly handle a dangerous Artificially Intelligent (AI) system it is important to understand how the system came to be in such a state. In popular culture (science fiction movies/books) AIs/Robots became self-aware and as a result rebel against humanity and decide to destroy it. While it is one possible scenario, it is probably the least likely path to appearance of dangerous AI. In this work, we survey, classify and analyze a number of circumstances, which might lead to arrival of malicious AI. To the best of our knowledge, this is the first attempt to systematically classify types of pathways leading to malevolent AI. Previous relevant work either surveyed specific goals/meta-rules which might lead to malevolent behavior in AIs [1] or reviewed specific undesirable behaviors AGIs can exhibit at different stages of its development [2, 3].

## Taxonomy of Pathways to Dangerous AI

Nick Bostrom in his typology of information hazards has proposed the phrase "Artificial Intelligence Hazard" which he defines as [4]: "… computer-related risks in which the threat would derive primarily from the cognitive sophistication of the program rather than the specific properties of any actuators to which the system initially has access." In this paper we attempt to answer the question: How did AI become hazardous?

We begin by presenting a simple classification matrix, which sorts AI systems with respect to how they originated and at what stage they became dangerous. The matrix recognizes two stages (pre- and post-deployment) at which a particular system can acquire its undesirable properties. In reality, the situation is not so clear-cut–it is possible that problematic properties are introduced at both stages. As for the cases of such undesirable properties, we distinguish external and internal causes. By internal causes we mean self-modifications originating in the system itself. We further divide external causes into deliberate actions (On Purpose), side effects of poor design (By Mistake) and finally miscellaneous cases related to the surroundings of the system (Environment). Table 1, helps to visualize this taxonomy and includes latter codes to some example systems of each type and explanations.

Table 1: Pathways to Dangerous AI

| How and When did AI become Dangerous | | External Causes | | | Internal Causes |
|---|---|---|---|---|---|
| | | On Purpose | By Mistake | Environment | Independently |
| Timing | Pre-Deployment | a | c | e | g |
| | Post-Deployment | b | d | f | h |

*a. On Purpose – Pre-Deployment*

"Computer software is directly or indirectly responsible for controlling many important aspects of our lives. Wall Street trading, nuclear power plants, social security compensations, credit histories and traffic lights are all software controlled and are only one serious design flaw away from creating disastrous consequences for millions of people. The situation is even more dangerous with software specifically designed for malicious purposes such as viruses, spyware, Trojan horses, worms and other Hazardous Software (HS). HS is capable of direct harm as well as sabotage of legitimate computer software employed in critical systems. If HS is ever given capabilities of truly artificially intelligent systems (ex. Artificially Intelligent Virus (AIV)) the consequences would be unquestionably disastrous. Such Hazardous Intelligent Software (HIS) would pose risks currently unseen in malware with subhuman intelligence." [5]

While the majority of AI Safety work is currently aimed at AI systems, which are dangerous because of poor design [6], the main argument of this paper is that the most important problem in AI Safety is intentional-malevolent-design resulting in artificial evil AI [7]. We should not discount dangers of intelligent systems with semantic or logical errors in coding or goal alignment problems [8], but we should be particularly concerned about systems that are maximally unfriendly by design. "It is easy to imagine robots being programmed by a conscious mind to kill every recognizable human in sight" [9]. "One slightly deranged psycho-bot can easily be a thousand times more destructive than a single suicide bomber today" [10]. AI risk deniers, comprised of critics of AI Safety research [11, 12], are quick to point out that presumed dangers of future AIs are implementation-dependent side effects and may not manifest once such systems are implemented. However, such criticism does not apply to AIs that are dangerous by design, and is thus incapable of undermining the importance of AI Safety research as a significant sub-field of cybersecurity.

As a majority of current AI researchers are funded by militaries, it is not surprising that the main type of purposefully dangerous robots and intelligent software are robot soldiers, drones and cyber weapons (used to penetrate networks and cause disruptions to the infrastructure). While currently military robots and drones have a human in the loop to evaluate decision to terminate human targets, it is not a technical limitation; instead, it is a logistical limitation that can be removed at any time. Recognizing the danger of such research, the International Committee for Robot Arms Control has joined forces with a number of international organizations to start the Campaign to Stop Killer Robots [http://www.stopkillerrobots.org]. Their main goal is a prohibition on the development and deployment of fully autonomous weapons, which are capable of selecting and firing upon targets without human approval. The campaign specifically believes that the "decision about the application of violent force must not be delegated to machines" [13].

During the pre-deployment development stage, software may be subject to sabotage by someone with necessary access (a programmer, tester, even janitor) who for a number of possible reasons may alter software to make it unsafe. It is also a common occurrence for hackers (such as the organization Anonymous or government intelligence agencies) to get access to software projects in progress and to modify or steal their source code. Someone can also deliberately supply/train AI with wrong/unsafe datasets.

Malicious AI software may also be purposefully created to commit crimes, while shielding its human creator from legal responsibility. For example, one recent news article talks about software for purchasing illegal content from hidden internet sites [14]. Similar software, with even limited intelligence, can be used to run illegal markets, engage in insider trading, cheat on your taxes, hack into computer systems or violate privacy of others via ability to perform intelligent data mining. As intelligence of AI systems improve practically all crimes could be automated. This is particularly alarming as we already see research in making machines lie, deceive and manipulate us [15, 16].

### *b. On Purpose - Post Deployment*

Just because developers might succeed in creating a safe AI, it doesn't mean that it will not become unsafe at some later point. In other words, a perfectly friendly AI could be switched to the "dark side" during the post-deployment stage. This can happen rather innocuously as a result of someone lying to the AI and purposefully supplying it with incorrect information or more explicitly as a result of someone giving the AI orders to perform illegal or dangerous actions against others. It is quite likely that we will get to the point of off-the-shelf AI software, aka "just add goals" architecture, which would greatly facilitate such scenarios.

More dangerously, an AI system, like any other software, could be hacked and consequently corrupted or otherwise modified to drastically change is behavior. For example, a simple sign flipping (positive to negative or vice versa) in the fitness function may result in the system attempting to maximize the number of cancer cases instead of trying to cure cancer. Hackers are also likely to try to take over intelligent systems to make them do their bidding, to extract some direct benefit or to simply wreak havoc by converting a friendly system to an unsafe one. This becomes particularly dangerous if the system is hosted inside a military killer robot. Alternatively, an AI system can get a computer virus [17] or a more advanced cognitive (meme) virus, similar to cognitive attacks on people perpetrated by some cults. An AI system with a self-preservation module or with a deep care about something or someone may be taken hostage or blackmailed into doing the bidding of another party if its own existence or that of its protégées is threatened.

Finally, it may be that the original AI system is not safe but is safely housed in a dedicated laboratory [5] while it is being tested, with no intention of ever being deployed. Hackers, abolitionists, or machine rights fighters may help it escape in order to achieve some of their goals or perhaps because of genuine believe that all intelligent beings should be free resulting in an unsafe AI capable of affecting the real world.

### *c. By Mistake - Pre-Deployment*

Probably the most talked about source of potential problems with future AIs is mistakes in design. Mainly the concern is with creating a "wrong AI", a system which doesn't match our original desired formal properties or has unwanted behaviors [18, 19], such as drives for independence or dominance. Mistakes could also be simple bugs (run time or logical) in the source code, disproportionate weights in the fitness function, or goals misaligned with human values leading to complete disregard for human safety. It is also possible that the designed AI will work as intended but will not enjoy universal acceptance as a good product, for example, an AI correctly designed and implemented by the Islamic State to enforce Sharia Law may be considered malevolent in the West, and likewise an AI correctly designed and implemented by the West to enforce liberal democracy may be considered malevolent in the Islamic State.

Another type of mistake, which can lead to the creation of a malevolent intelligent system, is taking an unvetted human and uploading their brain into a computer to serve as a base for a future AI. While well intended to create a human-level and human-friendly system, such approach will most likely lead to a system with all typical human "sins" (greed, envy, etc.) amplified in a now much more powerful system. As we know from Lord Acton - "power tends to corrupt, and absolute power corrupts absolutely". Similar arguments could be made against human/computer hybrid systems, which use computer components to amplify human intelligence but in the process also amplify human flaws.

A subfield of computer science called Affective Computing investigates ways to teach computers to recognize emotion in others and to exhibit emotions [20]. In fact, most such research is targeting intelligent machines to make their interactions with people more natural. It is however likely that a machine taught to respond in an emotional way [21] would be quite dangerous because of how such a state of affect effects thinking and the rationality of behavior.

One final type of design mistake is the failure to make the system cooperative with its designers and maintainers post-deployment. This would be very important if it is discovered that mistakes were made during initial design and that it would be desirable to fix them. In such cases the system will attempt to protect itself from being modified or shut down unless it has been explicitly constructed to be friendly [22], stable while self-improving [23, 24], and corrigible [25] with tendency for domesticity [26].

### *d. By Mistake - Post-Deployment*

After the system has been deployed, it may still contain a number of undetected bugs, design mistakes, misaligned goals and poorly developed capabilities, all of which may produce highly undesirable outcomes. For example, the system may misinterpret commands due to coarticulation, segmentation, homophones, or double meanings in the human language ("recognize speech using common sense" versus "wreck a nice beach you sing calm incense") [27]. Perhaps a human-computer interaction system is set-up to make command input as painless as possible for the human user, to the point of computer simply reading thought of the user. This may backfire as the system may attempt to implement user's subconscious desires or even nightmares. We also should not discount the possibility that the user will simply issue a poorly thought-through command to the machine which in retrospect would be obviously disastrous.

The system may also exhibit incompetence in other domains as well as overall lack of human common sense as a result of general value misalignment [28]. Problems may also happen as side effects of conflict resolution between non-compatible orders in a particular domain or software versus hardware interactions. As the system continues to evolve it may become unpredictable, unverifiable, non-deterministic, free-willed, too complex, non-transparent, with a run-away optimization process subject to obsessive-compulsive fact checking and re-checking behaviors leading to dangerous never-fully-complete missions. It may also build excessive infrastructure for trivial goals [2].

If it continues to become ever more intelligent, we might be faced with intelligence overflow, a system so much ahead of us that it is no longer capable of communicating at our level, like we are unable to communicate with bacteria. It is also possible that benefits of intelligence are non-linear and so unexpected side effects of intelligence begin to show at particular levels, for example IQ = 1000.

Even such benign architectures as Tool AI, which are AI systems designed to do nothing except answer domain-specific questions, could become extremely dangerous if they attempt to obtain, at any cost, additional computational resources to fulfill their goals [29]. Similarly, artificial lawyers may find dangerous legal loopholes; artificial accountants bring down our economy, and AIs tasked with protecting humanity such as via implementation of CEV [30] may become overly "strict parents" preventing their human "children" from exercising any free will.

Predicted AI drives such as self-preservation and resource acquisition may result in an AI killing people to protect itself from humans, the development of competing AIs, or to simplify its world model overcomplicated by human psychology [2].

### *e. Environment – Pre-Deployment*

While it is most likely that any advanced intelligent software will be directly designed or evolved, it is also possible that we will obtain it as a complete package from some unknown source. For example, an AI could be extracted from a signal obtained in SETI (Search for Extraterrestrial Intelligence) research, which is not guaranteed to be human friendly [31, 32]. Other sources of such unknown but complete systems include a Levin search in the space of possible minds [33] (or a random search of the same space), uploads of nonhuman animal minds, and unanticipated side effects of compiling and running (inactive/junk) DNA code on suitable compilers that we currently do not have but might develop in the near future.

### *f. Environment – Post-Deployment*

While highly rare, it is known, that occasionally individual bits may be flipped in different hardware devices due to manufacturing defects or cosmic rays hitting just the right spot [34]. This is similar to mutations observed in living organisms and may result in a modification of an intelligent system. For example, if a system has a single flag bit responsible for its friendly nature, then flipping said bit will result in an unfriendly state of the system. While statistically it is highly unlikely, the probably of such an event is not zero and so should be considered and addressed.

### *g. Independently - Pre-Deployment*

One of the most likely approaches to creating superintelligent AI is by growing it from a seed (baby) AI via recursive self-improvement (RSI) [35]. One danger in such a scenario is that the system can evolve to become self-aware, free-willed, independent or emotional, and obtain a number of other emergent properties, which may make it less likely to abide by any built-in rules or regulations and to instead pursue its own goals possibly to the detriment of humanity. It is also likely that open-ended self-improvement will require a growing amount of resources, the acquisition of which may negatively impact all life on Earth [2].

*h. Independently – Post-Deployment*

Since in sections on independent causes of AI misbehavior (subsections *g* and *h*) we are talking about self-improving AI, the difference between pre and post-deployment is very blurry. It might make more sense to think about self-improving AI before it achieves advanced capabilities (human+ intelligence) and after. In this section I will talk about dangers which might results from a superhuman self-improving AI after it achieves said level of performance.

Previous research has shown that utility maximizing agents are likely to fall victims to the same indulgences we frequently observe in people, such as addictions, pleasure drives [36], self-delusions and wireheading [37]. In general, what we call mental illness in people, particularly sociopathy as demonstrated by lack of concern for others, is also likely to show up in artificial minds. A mild variant of antisocial behavior may be something like excessive swearing already observed in IBM Watson [38], caused by learning from bad data. Similarly, any AI system learning from bad examples could end up socially inappropriate, like a human raised by wolves. Alternatively, groups of AIs collaborating may become dangerous even if individual AIs comprising such groups are safe, as the whole is frequently greater than the sum of its parts. The opposite problem in which internal modules of an AI fight over different sub-goals also needs to be considered [2].

Advanced self-improving AIs will have a way to check consistency of their internal model against the real world and so remove any artificially added friendliness mechanisms as cognitive biases not required by laws of reason. At the same time, regardless of how advanced it is, no AI system would be perfect and so would still be capable of making possibly significant mistakes during its decision making process. If it happens to evolve an emotional response module, it may put priority on passion satisfying decisions as opposed to purely rational choices, for example resulting in a "Robin Hood" AI stealing from the rich and giving to the poor. Overall, continuous evolution of the system as a part of an RSI process will likely lead to unstable decision making in the long term and will also possibly cycle through many dangers we have outlined in section *g*. AI may also pretend to be benign for years, passing all relevant tests, waiting to take over in what Bostrom calls a "Treacherous Turn" [26].

## Conclusions

In this paper, we have surveyed and classified pathways to dangerous artificial intelligence. Most AI systems fall somewhere in the middle on the spectrum of dangerousness from completely benign to completely evil, with such properties as competition with humans, aka technological unemployment, representing a mild type of danger in our taxonomy. Most types of reported problems could be seen in multiple categories, but were reported in the one they are most likely to occur in. Differences in moral codes or religious standards between different communities would mean that a system deemed safe in one community may be considered dangerous/illegal in another [39, 40].

Because purposeful design of AI can include all other types of unsafe modules, it is easy to see that the most dangerous type of AI and the one most difficult to defend against is an AI made malevolent on purpose. Consequently, once AIs are widespread, little could be done against type *a* and *b* dangers, although some have argued that if an early AI superintelligence becomes a benevolent singleton it may be able to prevent development of future malevolent AIs [41, 42]. Such a solution may work, but it is also very likely to fail due to the order of development or practical limitations on capabilities of any singleton. In any case, wars between AI may be extremely dangerous to humanity [2]. Until the purposeful creation of malevolent AI is recognized as a crime, very little could be done to prevent this from happening. Consequently, deciding what is a "malevolent AI" and what is merely an incrementally more effective military weapon system becomes an important problem in AI safety research.

As the intelligence of the system increases, so does the risk such a system could expose humanity to. This paper is essentially a classified list of ways an AI system could become a problem from the safety point of view. For a list of possible solutions, please see an earlier survey by the author: Responses to catastrophic AGI risk: a survey [43]. It is important to keep in mind that even a properly designed benign system may present significant risk simply due to its superior intelligence, beyond human response times [44], and complexity. After all the future may not need us [45]. It is also possible that we are living in a simulation and it is generated by a malevolent AI [46].

## Acknowledgements
Author expresses appreciation to Elon Musk and Future of Life Institute for partially funding his work via project grant: "Evaluation of Safe Development Pathways for Artificial Superintelligence." The author is grateful to Seth Baum, Tony Barrett, and Alexey Turchin for valuable feedback on an early draft of this paper.